\documentclass[runningheads,a4paper]{llncs}

\usepackage{amssymb}
\usepackage{amsmath}
\usepackage{graphicx}
\usepackage{color}
\usepackage[table,xcdraw]{xcolor}
\usepackage{url}
\usepackage{multirow}
\usepackage{hyperref}
\usepackage{subfig}
\usepackage{todonotes}
\usepackage[misc]{ifsym}

\urldef{\mail}\path|j.m.wolterink@umcutrecht.nl|

\newcommand{\keywords}[1]{\par\addvspace\baselineskip
\noindent\keywordname\enspace\ignorespaces#1}

\usepackage{array}
\newcolumntype{L}[1]{>{\raggedright\let\newline\\\arraybackslash\hspace{0pt}}m{#1}}
\newcolumntype{C}[1]{>{\centering\let\newline\\\arraybackslash\hspace{0pt}}m{#1}}
\newcolumntype{R}[1]{>{\raggedleft\let\newline\\\arraybackslash\hspace{0pt}}m{#1}}

\begin{document}

\mainmatter  

\title{Automatic Segmentation and Disease Classification Using Cardiac Cine MR Images}
\titlerunning{Automatic Segmentation and Diagnosis Using Cardiac Cine MR Images}

\author{Jelmer M. Wolterink\inst{1}$^{\text{(\Letter)}}$, Tim Leiner\inst{2}, Max A. Viergever\inst{1}, Ivana I\v{s}gum\inst{1}}

\authorrunning{J.M. Wolterink et al.}

\institute{Image Sciences Institute, University Medical Center Utrecht, The Netherlands \\ \mail \and Department of Radiology, University Medical Center Utrecht, The Netherlands}
\maketitle

\begin{abstract}
Segmentation of the heart in cardiac cine MR is clinically used to quantify cardiac function. We propose a fully automatic method for segmentation and disease classification using cardiac cine MR images.

A convolutional neural network (CNN) was designed to simultaneously segment the left ventricle (LV), right ventricle (RV) and myocardium in end-diastole (ED) and end-systole (ES) images. Features derived from the obtained segmentations were used in a Random Forest classifier to label patients as suffering from dilated cardiomyopathy, hypertrophic cardiomyopathy, heart failure following myocardial infarction, right ventricular abnormality, or no cardiac disease.

The method was developed and evaluated using a balanced dataset containing images of 100 patients, which was provided in the MICCAI 2017 automated cardiac diagnosis challenge (ACDC). Segmentation and classification pipeline were evaluated in a four-fold stratified cross-validation. Average Dice scores between reference and automatically obtained segmentations were 0.94, 0.88 and 0.87 for the LV, RV and myocardium. The classifier assigned 91\% of patients to the correct disease category. Segmentation and disease classification took 5 s per patient.

The results of our study suggest that image-based diagnosis using cine MR cardiac scans can be performed automatically with high accuracy.
\end{abstract}

\section{Introduction}
Quantification of volumetric changes in the heart during the cardiac cycle is essential for diagnosis and monitoring of cardiac diseases. To this end, quantitative indices such as the ejection fraction and myocardial mass are typically extracted based on segmentations of the ventricular cavities and myocardium and used to identify patients suffering from cardiac diseases \cite{Marc10}.

However, segmentation of the ventricular cavities and myocardium in cine MR is a challenging problem \cite{Peti11}. Cine MR images are highly anisotropic, contrast in these images may be poor, and cardiac diseases may cause large variations in patient anatomy.  
The development of accurate cine MR segmentations methods is an ongoing endeavor \cite{Peti15}, which has recently seen contributions from deep learning methods, e.g. \cite{Tran16,Liem17}.

We propose a method for fully automatic segmentation of the LV cavity, the RV cavity and the myocardium in cardiac cine MR images. We use a deep learning method for cardiac cine MR segmentation and show that this method achieves high overlap with manual reference segmentations. Furthermore, we show how basic quantitative features extracted from the automatically obtained segmentations can be combined with patient information in a forest of randomized decision trees. This allows fast and accurate disease classification in cardiac patients, and detection of patients with ambiguous indications.

\section{Data}
The proposed method was developed and evaluated using data from the MICCAI 2017 \textit{a}utomated \textit{c}ardiac \textit{d}iagnosis \textit{c}hallenge (ACDC). This challenge provides a dataset consisting of cine MR images of 150 patients who have been clinically diagnosed in five classes: normal, dilated cardiomyopathy (DCM), hypertrophic cardiomyopathy (HCM), heart failure with infarction (MINF), or right ventricular abnormality (RVA). Thirty cases are provided in each class. 
The data set was separated by the challenge organizers into 100 training cases for which a reference standard was provided, and 50 test cases for which no reference standard was provided. Here, we describe experiments and results using the 100 training cases. 

For each patient, short axis (SA) cine MR images with 12-35 frames are available, in which the end-diastole (ED) and end-systole (ES) frame have been indicated. The image slices cover the LV from the base to the apex. In-plane voxel spacing varies from 1.37 to 1.68 mm and inter-slice spacing varies from 5 to 10 mm. Manual reference segmentations of the LV cavity, RV cavity and myocardium at ED and ES are provided. 

To correct for differences in voxel size, all 2D image slices were resampled to $1.4\times 1.4$ mm$^2$ spacing. Furthermore, to correct for image intensity differences between images, each MR volume was normalized between $[0.0,1.0]$ according to the 5th and 95th percentile of intensities in the image. 

\section{Methods}
We propose a fully automatic method for segmentation and diagnosis in cardiac cine MR images. The method uses a convolutional neural network (CNN) to segment the LV cavity, the RV cavity, and the myocardium in 2D short-axis cine MR slices. Quantitative indices are extracted from the obtained segmentations and combined with patient information in a Random Forest classifier that assigns patients to one of five classes (normal, DCM, HCM, MINF, RVA).

\begin{figure}[tp]
\centering
\includegraphics[width=0.9\textwidth]{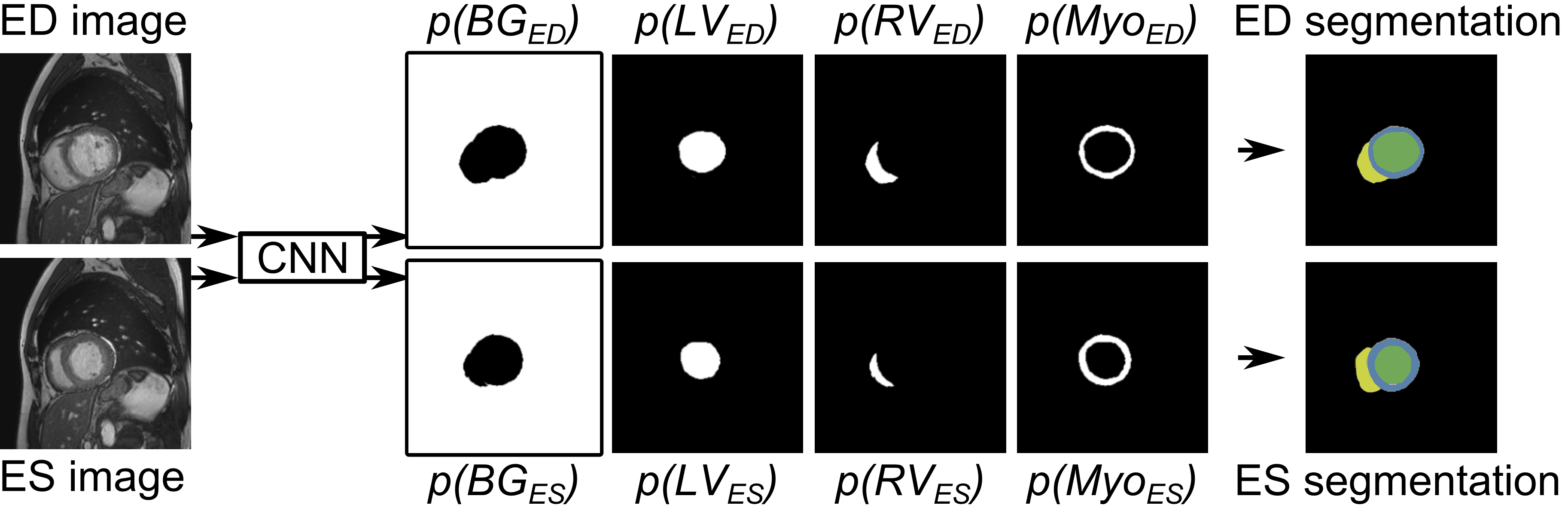}
\caption{Convolutional neural network for segmentation. The CNN uses anatomically aligned end-diastole (ED) and end-systole (ES) 2D image slices as input and simultaneously predicts probability maps for the background (BG), LV cavity (LV), RV cavity (RV) and myocardium (Myo) at ED and ES. These are combined into two multi-class segmentations.}
\label{fig:scheme}
\end{figure}

\subsection{Segmentation}
A CNN was trained to segment the LV cavity, the RV cavity, and the myocardium in 2D short-axis cine MR slices. Motivated by \cite{Yu15,Wolt16b}, the network was designed to contain a number of convolutional layers with increasing levels of dilation. This ensures a large receptive field with few trainable parameters and high resolution feature maps. The final receptive field for each voxel was $131\times 131$ voxels, or $18.3\times 18.3$ cm$^2$ in the resampled 2D slices. Potential overfitting of the network was mitigated by the inclusion of Batch Normalization layers \cite{Ioff15}.

Cine cardiac MR slices obtained throughout the cardiac cycle are anatomically aligned, but cardiac motion causes differences between images at different time points. These differences are more pronounced in the heart than in other areas \cite{Ateh16}. We allowed the CNN to leverage this information for heart localization by simultaneously providing anatomically corresponding ED and ES slices in two input channels (Fig. \ref{fig:scheme}). The CNN had eight output channels; four for ED labels ($BG_{ED}$, $LV_{ED}$, $RV_{ED}$, $Myo_{ED}$),  and four for ES labels ($BG_{ES}$, $LV_{ES}$, $RV_{ES}$, $Myo_{ES}$). ED and ES output channels were separately normalized through softmax functions.
Hence, for both the ED and the ES image there were four probability maps summing to 1. A segmentation was obtained for both images by assigning the class with the highest probability to each voxel.

The trainable parameters in the CNN were optimized using a loss function based on the Dice similarity coefficient \cite{Mill16}. This partly corrects for class imbalance in the voxel labels. 
A soft Dice loss was used,

\begin{equation}
Dice_c = \frac{\sum_i^N R_c(i)A_c(i)}{\sum_i^N R_c(i) + \sum_i^N A_c(i)},
\end{equation}

where $R_c$ is the binary reference image for class $c$, $A_c$ is the probability map for class $c$, $N$ is the number of voxels, and $Dice_c$ is the Dice coefficient for class $c$. This coefficient was computed for all eight classes ($BG_{ED}$, $RV_{ED}$, $Myo_{ED}$, $LV_{ED}$, $BG_{ES}$, $RV_{ES}$, $Myo_{ES}$, $LV_{ES}$) and averaged to ensure joint optimization for all classes. 

The combination of multiple trained CNN models in an ensemble typically results in more accurate predictions, but at the cost of repeated training. To obtain multiple models with a single training phase, we used the snapshot ensemble technique proposed in \cite{Huan17}. Hence, the learning rate followed a cyclic scheme according to the equation 

\begin{equation}
\alpha_i = \frac{\alpha_0}{2}\bigg(\cos\bigg(\frac{\pi \text{mod} (t-1,M)}{M}\bigg)+1\bigg),
\end{equation} 

where $\alpha_i$ is the current learning rate, $\alpha_0$ is the initial learning rate and $M$ is the cycle length, i.e. the number of iterations before a reset of the learning rate to the initial value. We set the total number of iterations to 150,000 and reset the learning rate to $\alpha_0=0.2$ after every $M=10,000$ iterations. A copy of the model was stored before each learning rate reset. Stochastic gradient descent was used for training, with $L2$-regularization on the parameters of the CNN. In each iteration, the network was optimized with a mini-batch containing 4 images with $151\times 151$ voxel samples, padded to $281\times 281$ to accommodate the $131\times 131$ voxel receptive field. The training data was augmented by 90 degree rotations of the images and reference segmentations. 

During testing, pairs of ED and ES images were processed by the six stored versions of the model between 100,000 and 150,000 iterations. The six predicted probability maps for each class were averaged before the class label with the largest probability was assigned to each voxel. No post-processing was applied other than selection of the largest 3D 6-connected component for each class.

\subsection{Diagnosis}
Each patient was described by patient and image characteristics. Patient characteristics were patient weight (in kg) and patient height (in cm). Image characteristics were extracted from the automatically obtained segmentations: LV, RV and myocardial volume at ED and ES (in ml), the LV and RV ejection fraction (EF), the ratio between RV and LV volume at ED and ES, and the ratio between myocardial and LV volume at ED and ES. Hence, 14 features were used in total: 2 patient-based and 12 image-based features.

A five-class (normal, HCM, DCM, MINF, RVA) Random Forest classifier \cite{Brei01} was trained, consisting of 1,000 decision trees that were grown to full depth. For each case, a posterior probability distribution was obtained. Patients were assigned to the class with the highest probability, and the entropy in the probability distribution was determined to estimate uncertainty of the classifier.

\section{Experiments and Results}
The proposed method was evaluated using the ACDC training set in a stratified four-fold cross-validation experiment. For each fold the system was trained using 15 training patients from each of the five classes and evaluated using five test patients from each of the five classes. Quantitative indices used to train the Random Forest in a fold were obtained with the trained CNN for that fold. Hence, training and validation set were completely separated throughout both stages. We here present combined results on all 100 training images.

\begin{figure}[tp]
\centering
\includegraphics[width=0.15\textwidth]{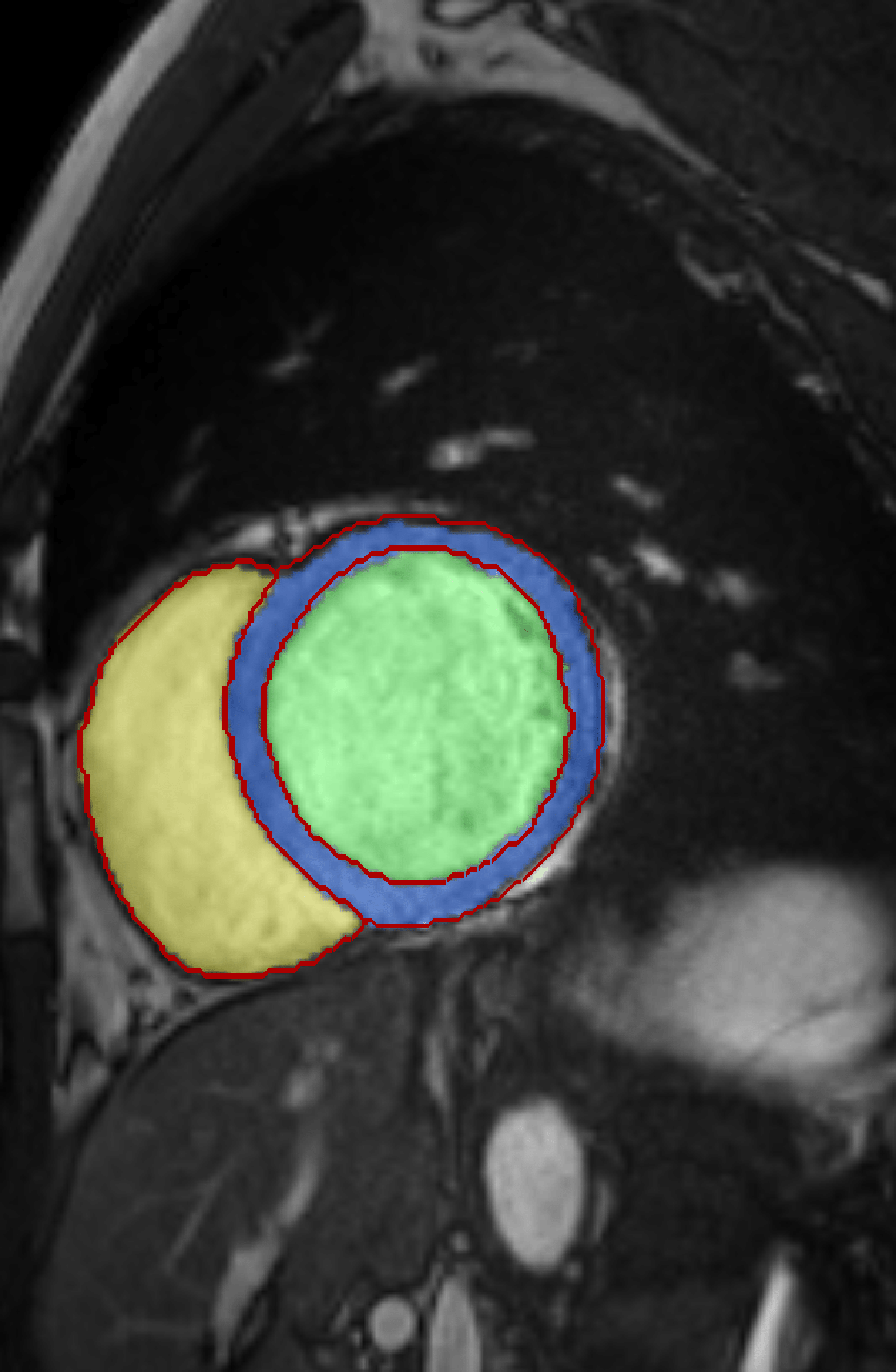}
\includegraphics[width=0.15\textwidth]{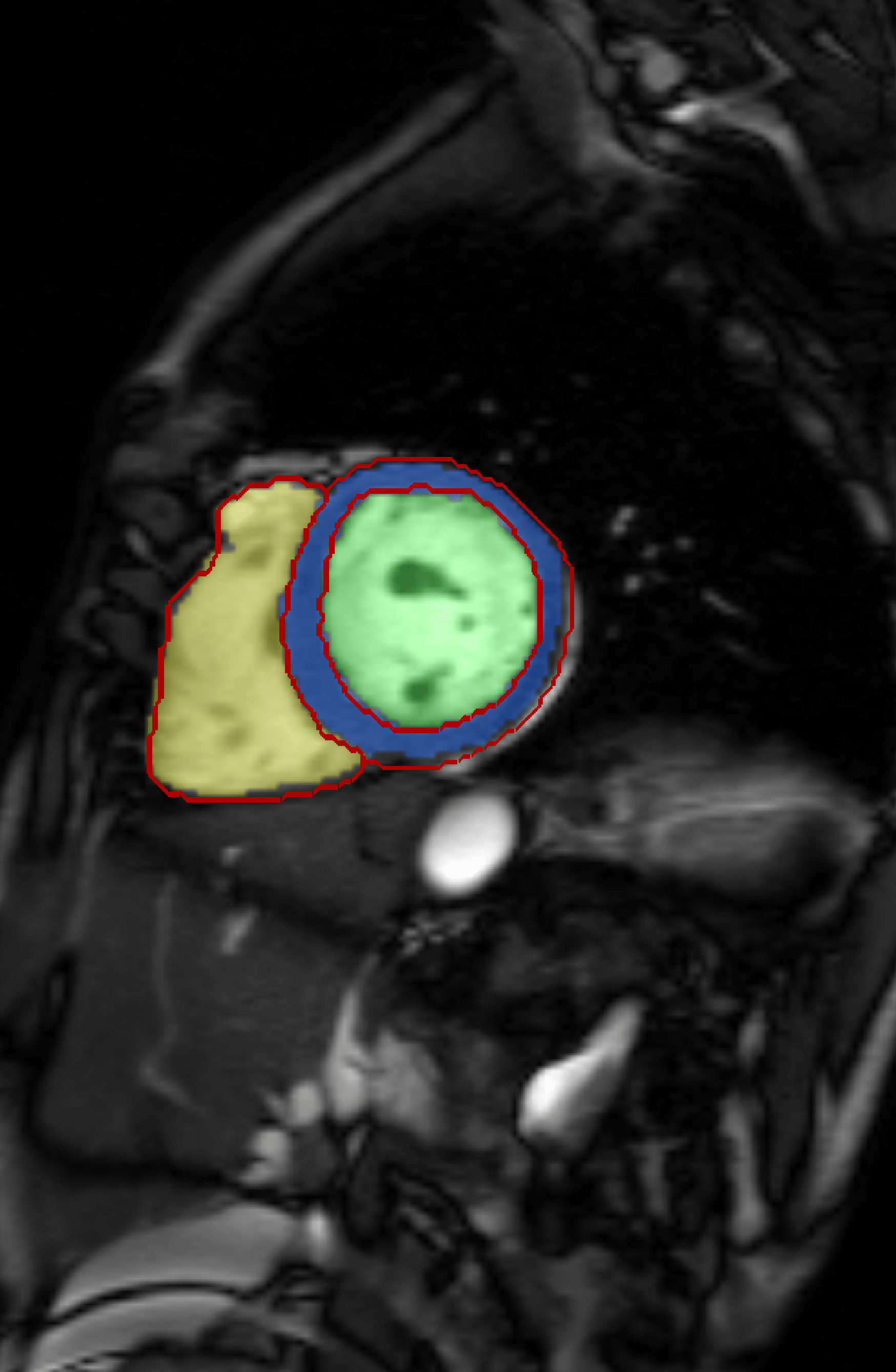}
\includegraphics[width=0.15\textwidth]{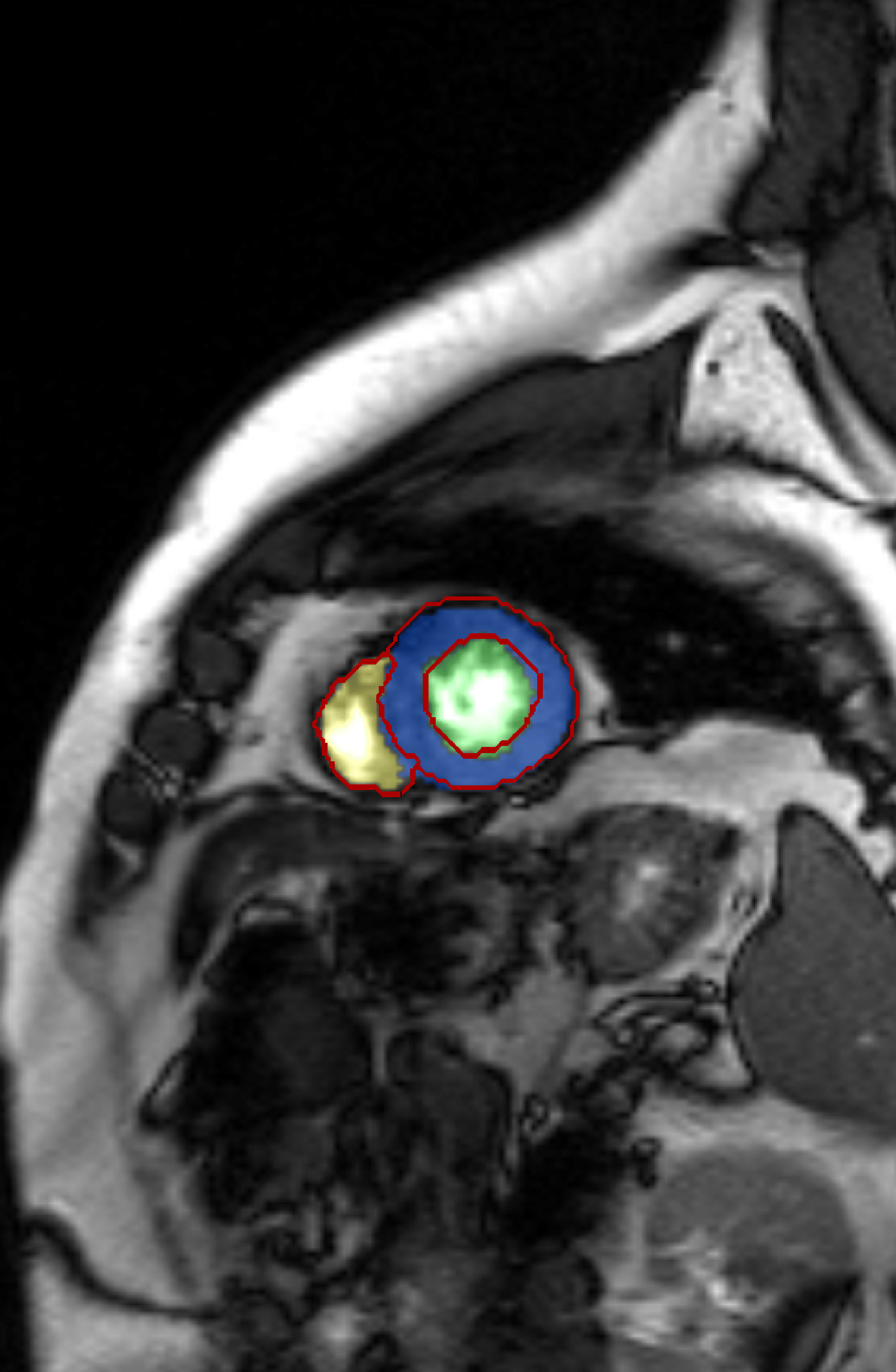}
\includegraphics[width=0.15\textwidth]{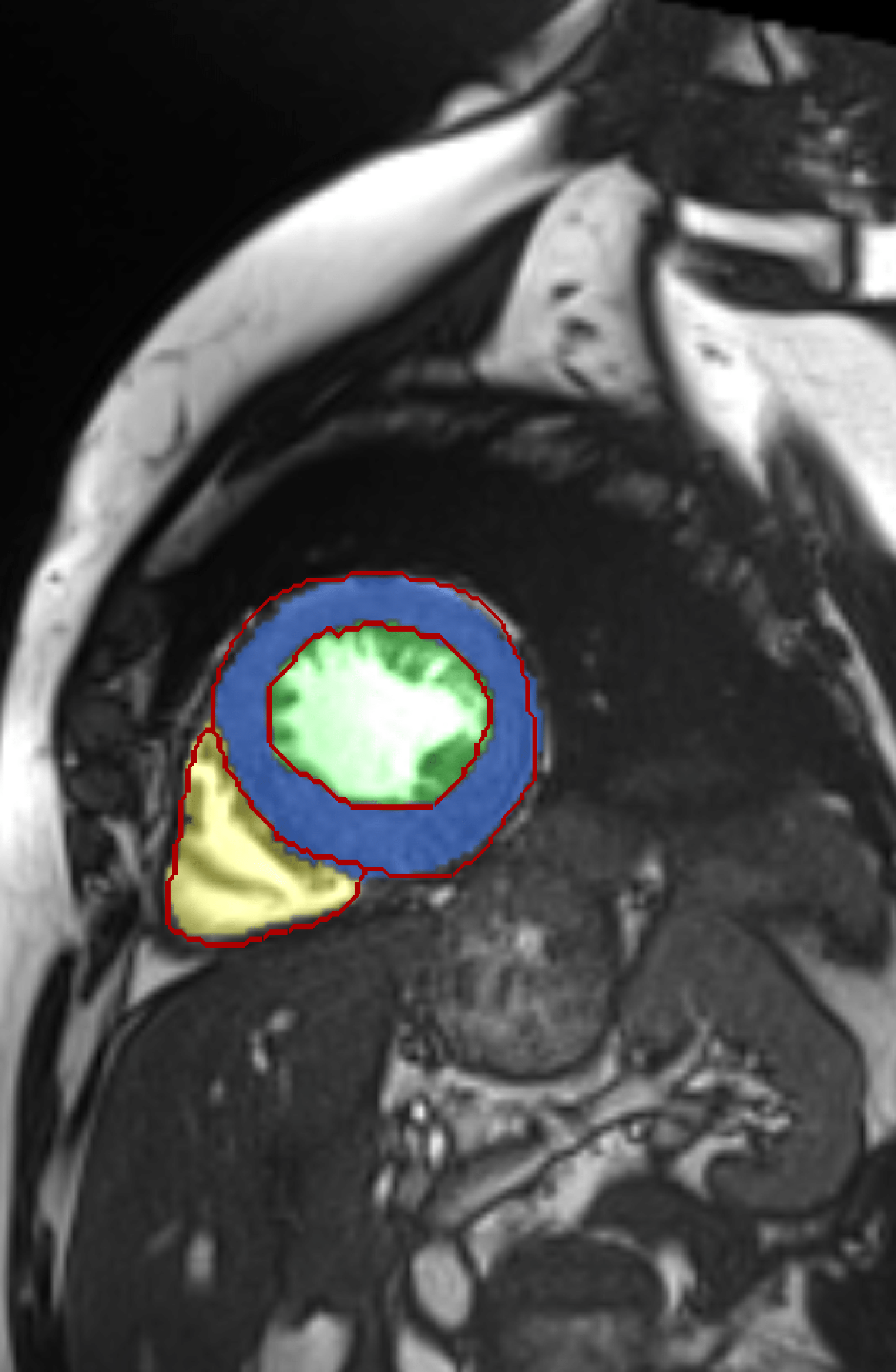}
\includegraphics[width=0.15\textwidth]{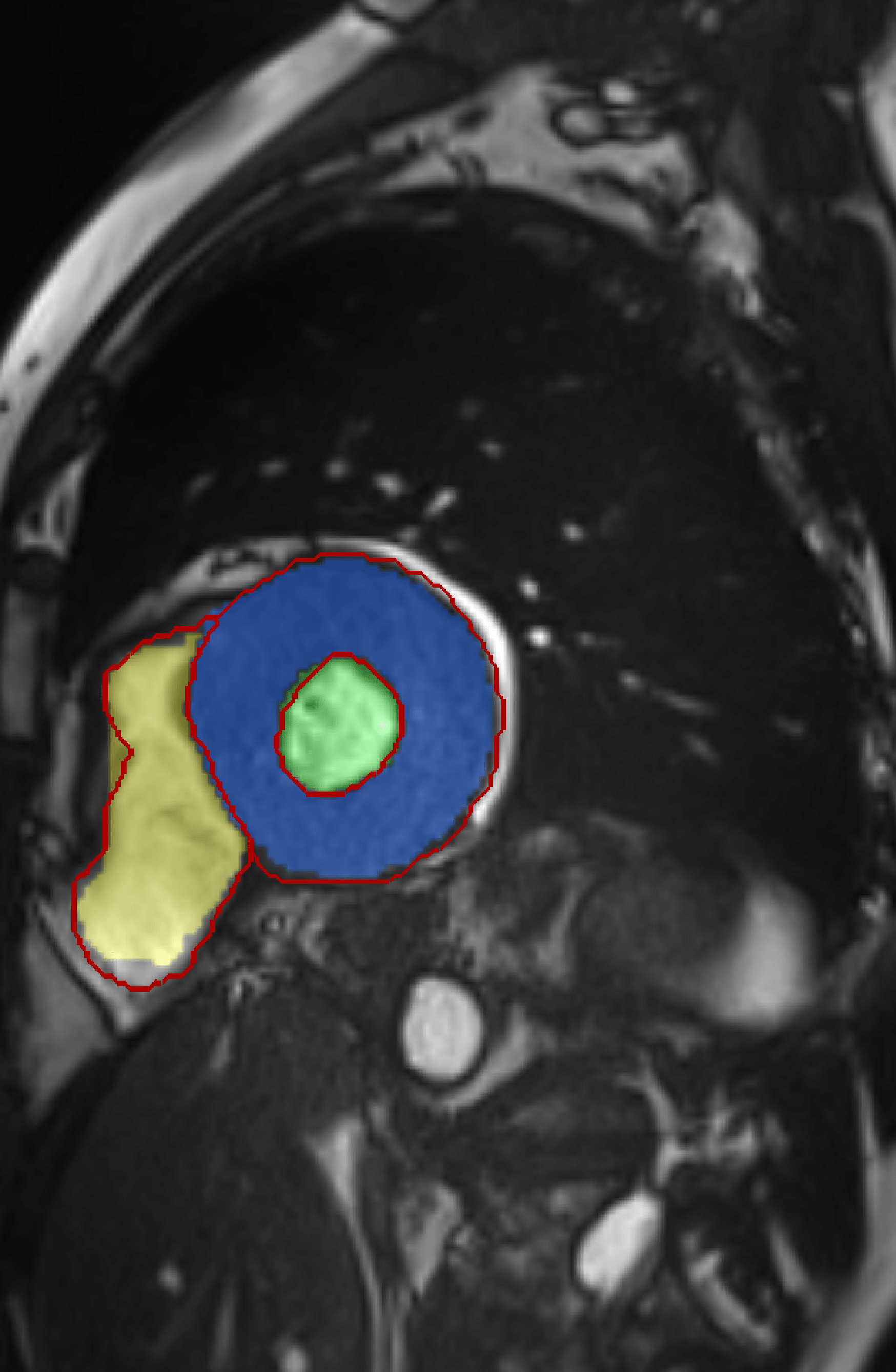}
\includegraphics[width=0.15\textwidth]{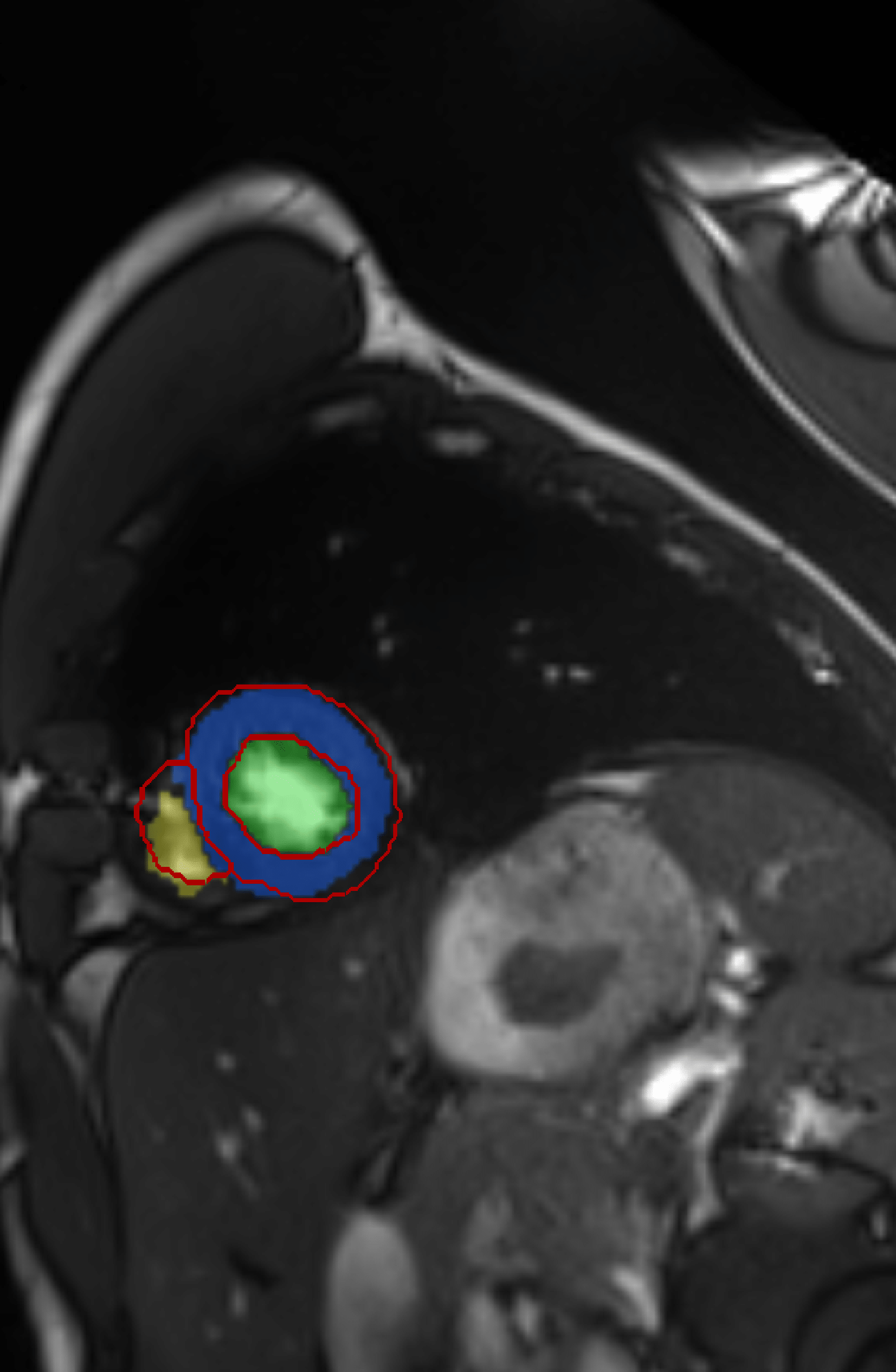}
\caption{Example segmentations obtained by the CNN in six different patients, showing the LV cavity in green, the RV cavity in yellow, and the myocardium in blue. Reference delineations are shown in red.}
\label{fig:examplesegs}
\end{figure}

\begin{table}[tp!]
\centering
\caption{Average Dice coefficients (Dice) and Hausdorff distances (HD, in mm) for LV, RV and myocardium segmentation at ED and ES.}
\label{tab:segresults}
\begin{tabular}{l|ll|ll|ll}
      & \multicolumn{2}{l|}{LV}         & \multicolumn{2}{l|}{RV}          & \multicolumn{2}{l}{Myocardium}  \\
      & Dice           & HD             & Dice           & HD              & Dice           & HD              \\ \hline
ED    & $0.96\pm 0.02$~~ & $8.35\pm 4.63$~~ & $0.92\pm 0.04$~~ & $13.39\pm 5.68$~~ & $0.86\pm 0.04$~~ & $11.77\pm 6.38$~~ \\
ES    & $0.91\pm 0.07$ & $9.01\pm 4.39$ & $0.84\pm 0.09$ & $15.03\pm 6.30$ & $0.88\pm 0.04$ & $10.85\pm 4.74$ \\
Total & $0.93\pm 0.05$ & $8.68\pm 4.51$ & $0.88\pm 0.08$ & $14.21\pm 6.04$  & $0.87\pm 0.04$ & $11.31\pm 5.62$
\end{tabular}
\end{table}

\begin{figure}[tp]
\centering
\includegraphics[width=0.41\textwidth]{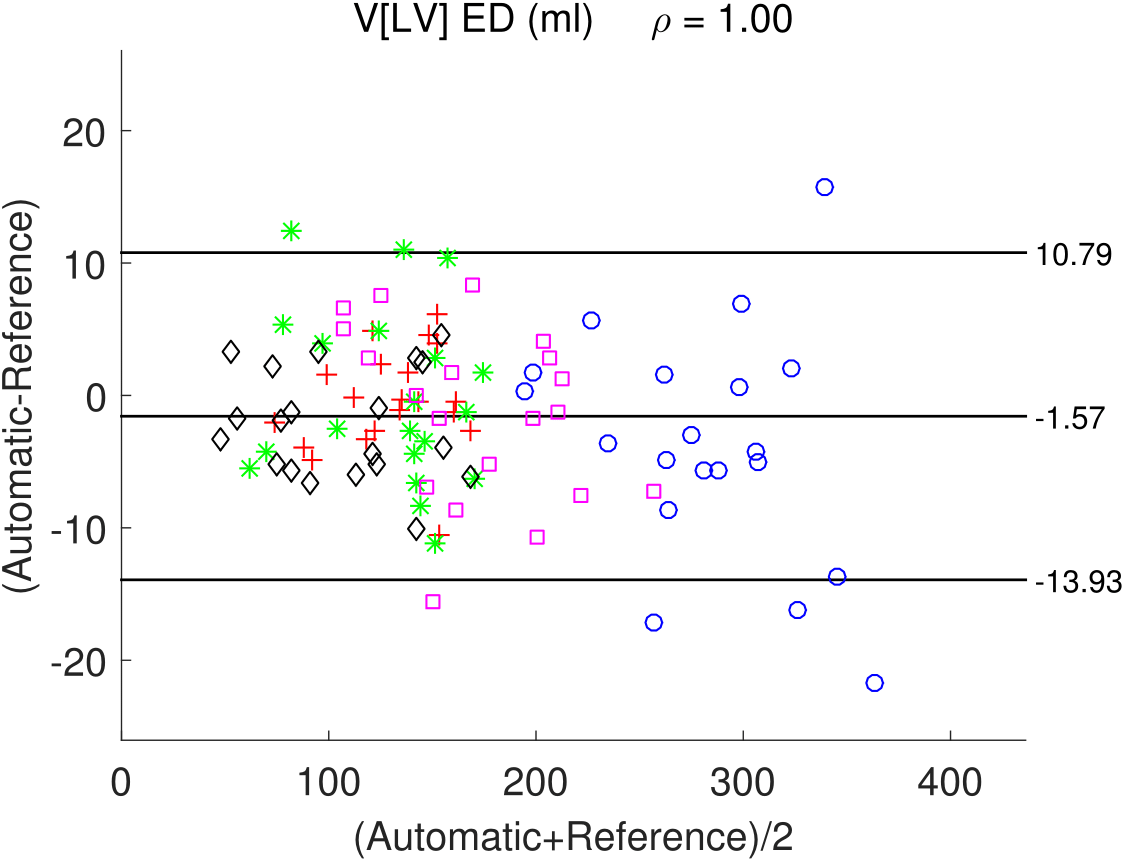}
\includegraphics[width=0.41\textwidth]{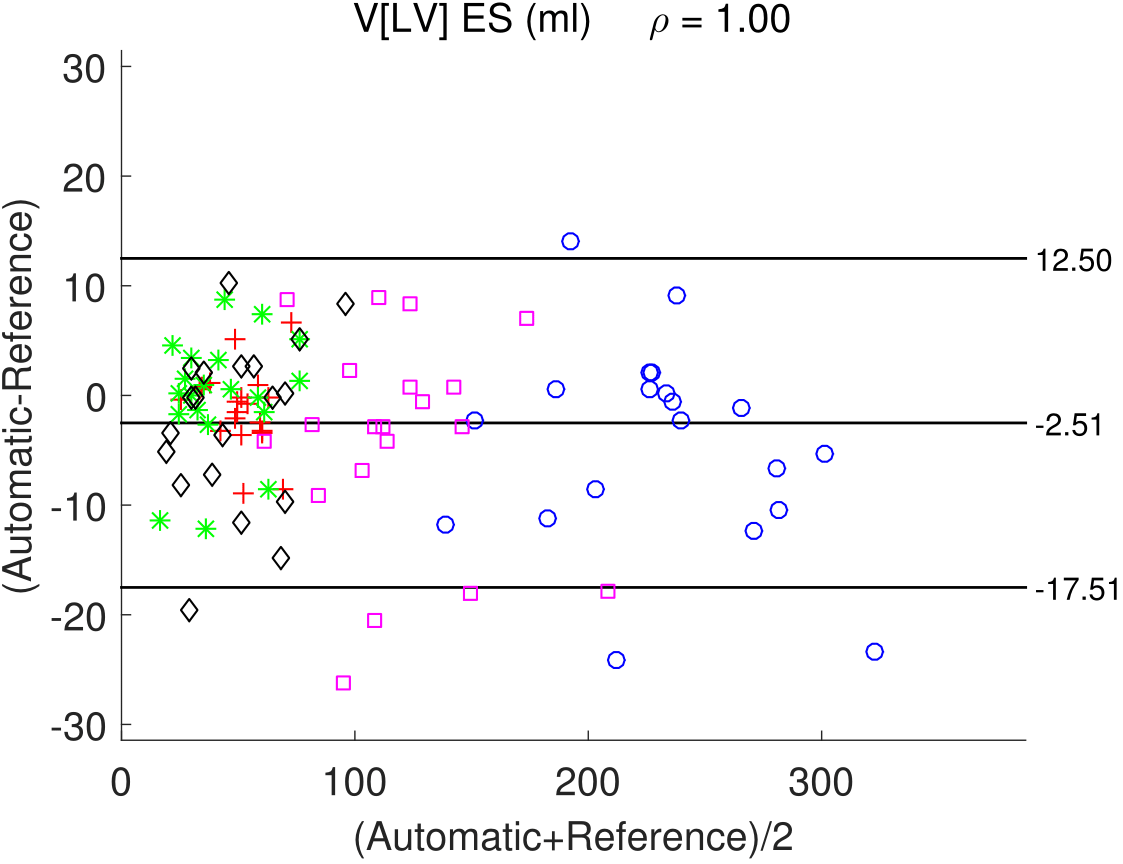} \\ \vspace{0.2cm}
\includegraphics[width=0.41\textwidth]{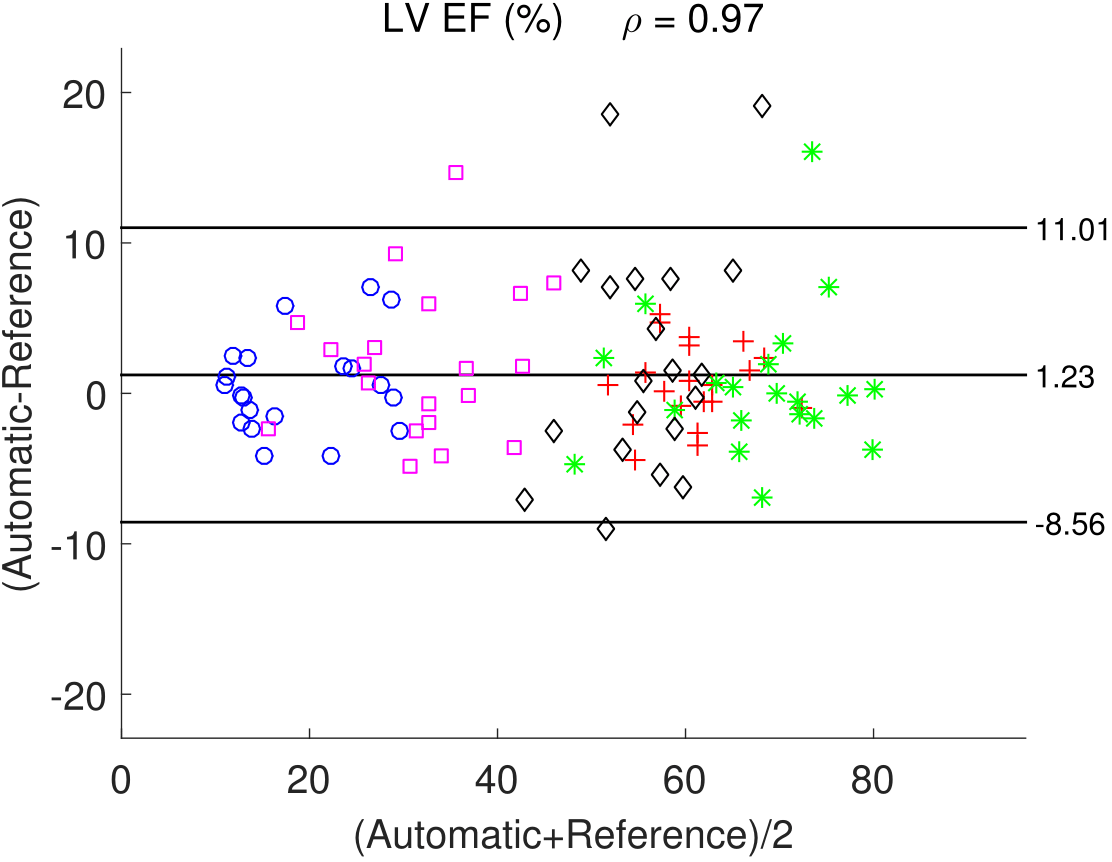}
\includegraphics[width=0.41\textwidth]{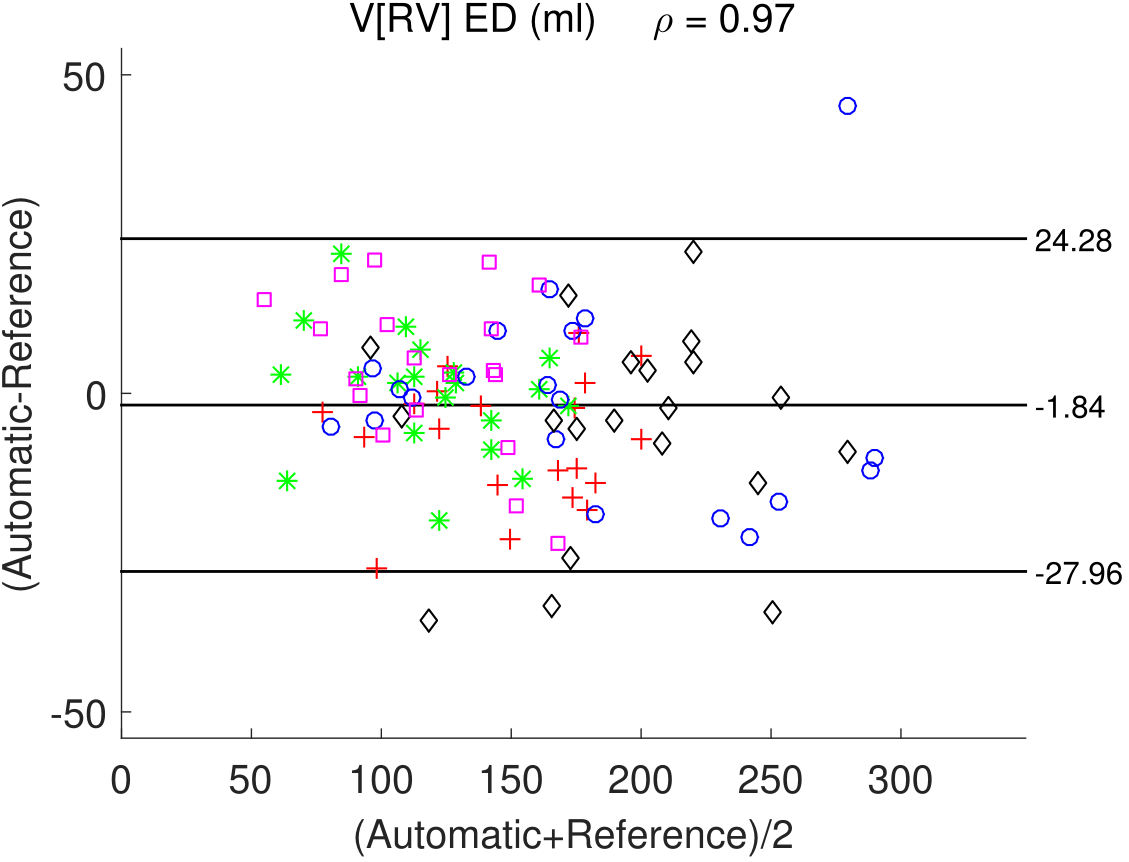} \\ \vspace{0.2cm}
\includegraphics[width=0.41\textwidth]{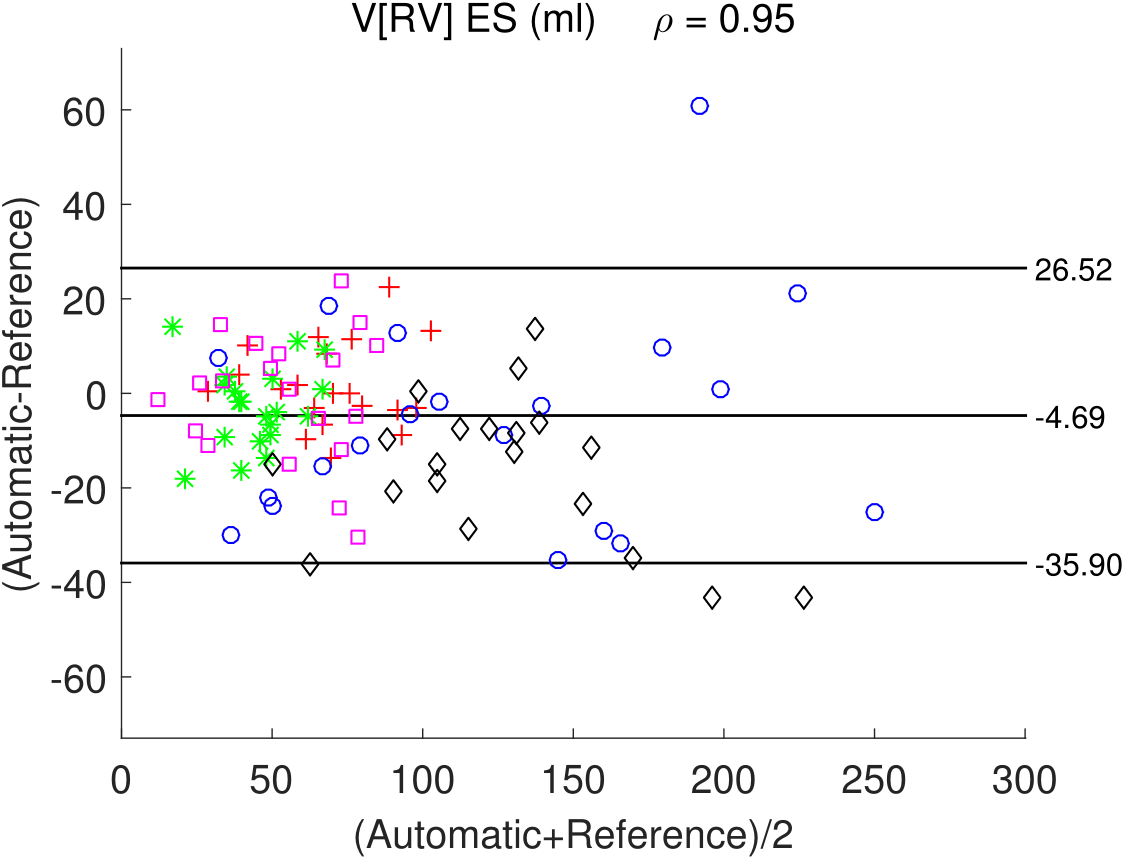}
\includegraphics[width=0.41\textwidth]{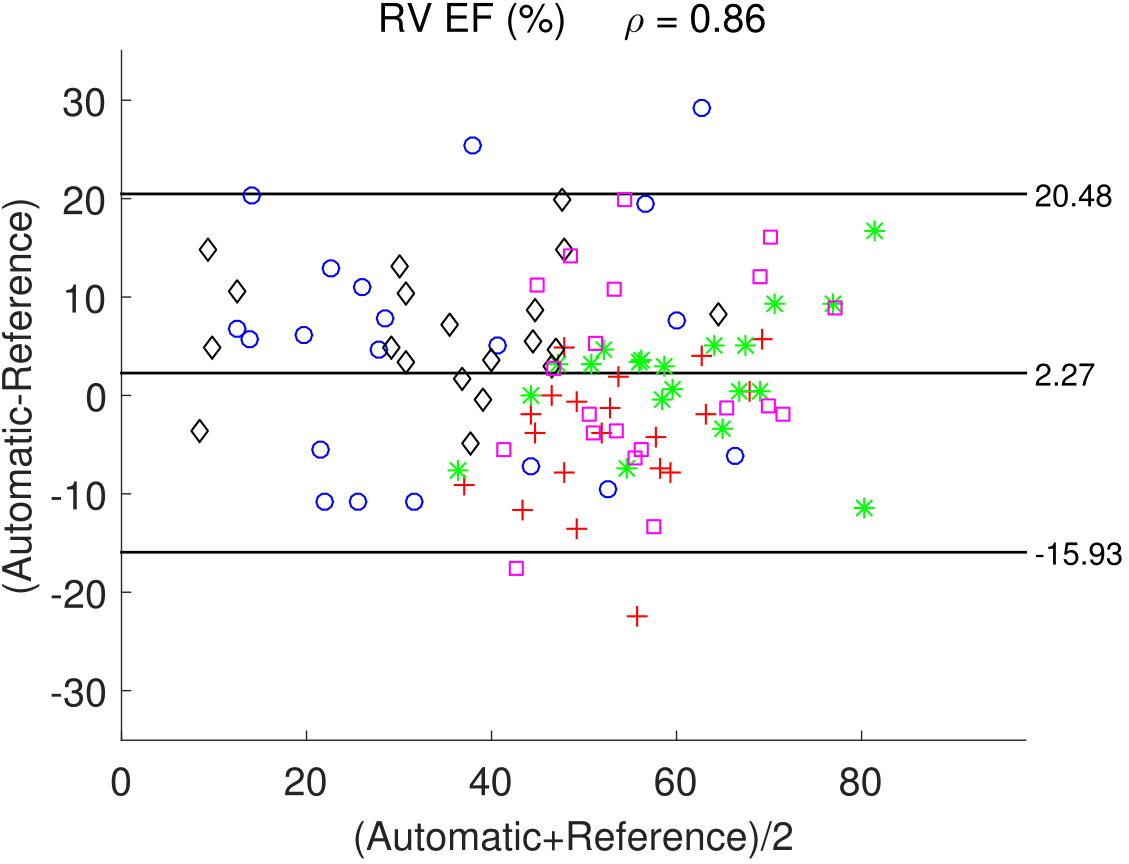} \\ \vspace{0.2cm}
\includegraphics[width=0.41\textwidth]{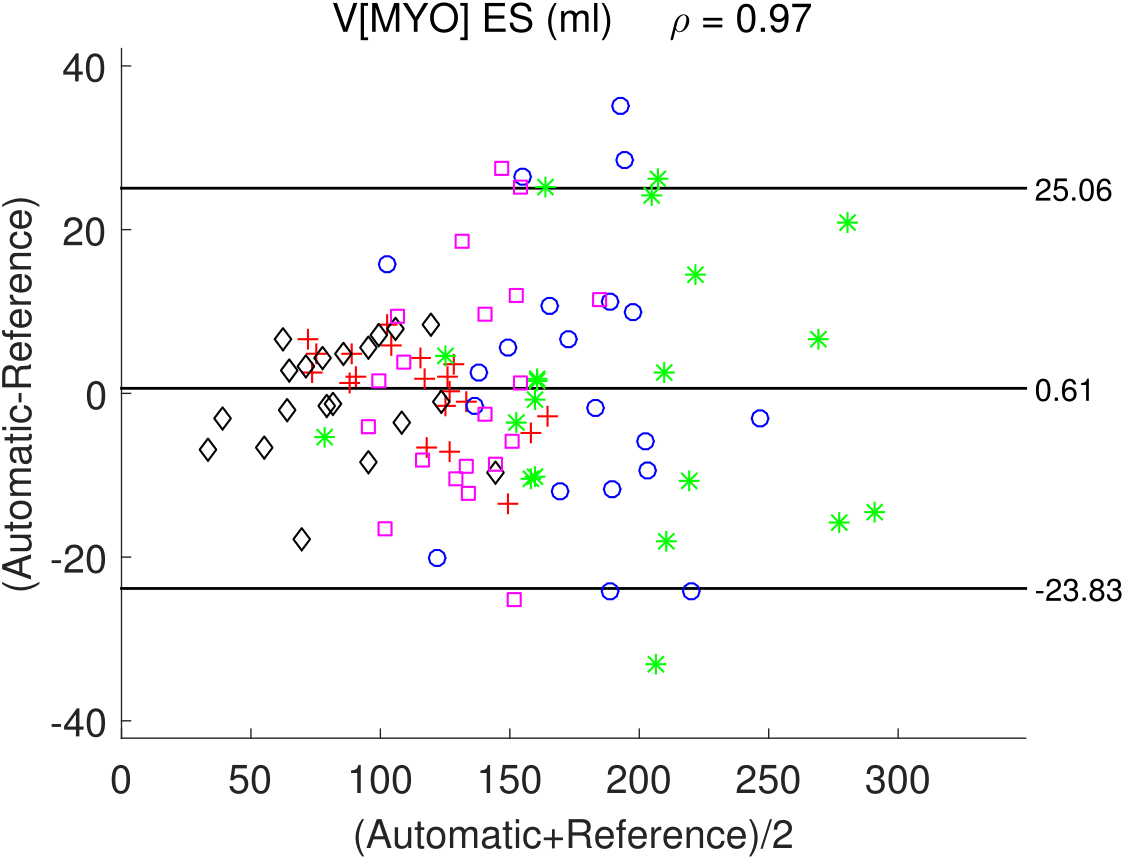}
\includegraphics[width=0.41\textwidth]{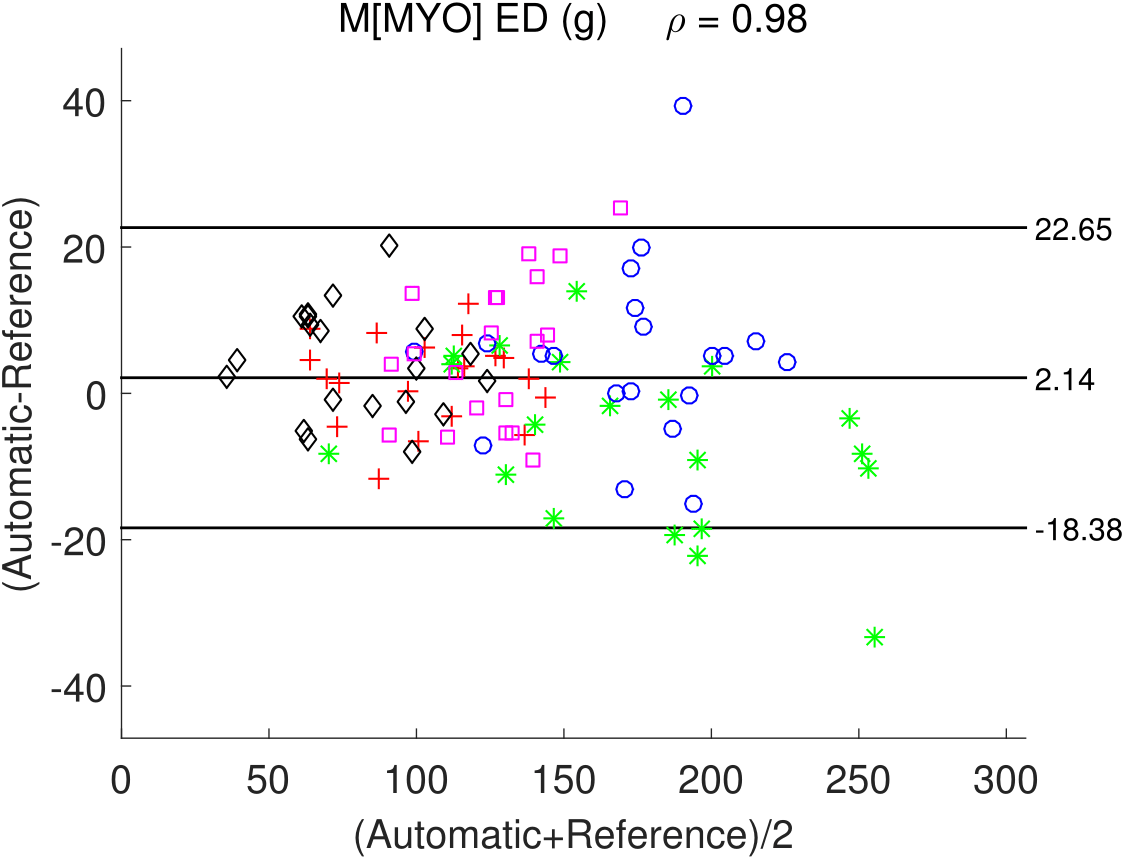} \\ \vspace{0.2cm}
\includegraphics[width=0.53\textwidth]{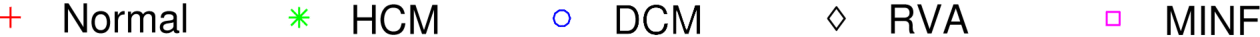}
\caption{Bland-Altman plots showing the agreement between reference and automatic quantification of the end-diastolic (ED) and end-systolic (ES) volume (V [in ml]) or mass (M [in g]) of the left ventricle (LV), right ventricle (RV) and myocardium (MYO). Bland-Altman limits of agreement and Pearson correlation values are listed. Points correspond to patients, with markers indicating classes.}
\label{fig:blandaltmans}
\end{figure}

\subsection{Segmentation Results}
Fig. \ref{fig:examplesegs} shows example segmentations and the corresponding reference delineations. All obtained segmentations were evaluated using the online platform provided by the organizers of the ACDC challenge. Table \ref{tab:segresults} shows Dice coefficients and Hausdorff distances for LV, RV and myocardium segmentation at ED and ES. Agreement with the reference standard was highest for the left ventricle at ED, and lowest for the right ventricle at ES. Performance was substantially higher in ED than in ES for the LV and RV, but not for the myocardium. Average Hausdorff distances were lower for the LV than for the RV and myocardium. This might be caused by the strong contrast that is typically present between the LV and the surrounding myocardium, and the poorer contrast between the myocardium and its surrounding structures. Furthermore, the shape of the RV is more irregular than that of the LV (Fig. \ref{fig:examplesegs}). 

In addition to the Dice coefficient and Hausdorff distance, the agreement between reference and automatically derived quantitative indices was determined using the ACDC challenge online platform. Fig. \ref{fig:blandaltmans} shows Bland-Altman plots with limits of agreement and Pearson correlations for the agreement between LV, RV and myocardium volume or mass quantification at ED and ES, as well as the LV and RV EF. There was a slight underestimation of the LV and RV at both ED and ES, and a slight overestimation of the myocardium at both ED and ES. The Pearson correlation between reference and automatically determined LV EF values was 0.97, while this correlation was 0.86 for the RV EF, reflecting lower segmentation accuracy for the RV. 

The CNN was implemented in Theano and Lasagne. Segmentation with an ensemble of six trained CNNs took 4 s per patient on a NVIDIA Titan X GPU.

\subsection{Diagnosis Results}
The obtained diagnoses were evaluated in the online platform provided by the ACDC organizers. Table \ref{tab:diagnosis} shows the confusion matrix for classification into five categories, with an overall accuracy of 91\%. Sensitivity was 100\% for the normal class, 90\% for the DCM, HCM and RVA classes, and 85\% for the MINF class. Four out of nine errors were made by confusion between myocardial infarction and dilated cardiomyopathy, both of which are characterized by low LV EF values. 

\begin{figure}[tp]
\centering
\includegraphics[width=0.32\textwidth]{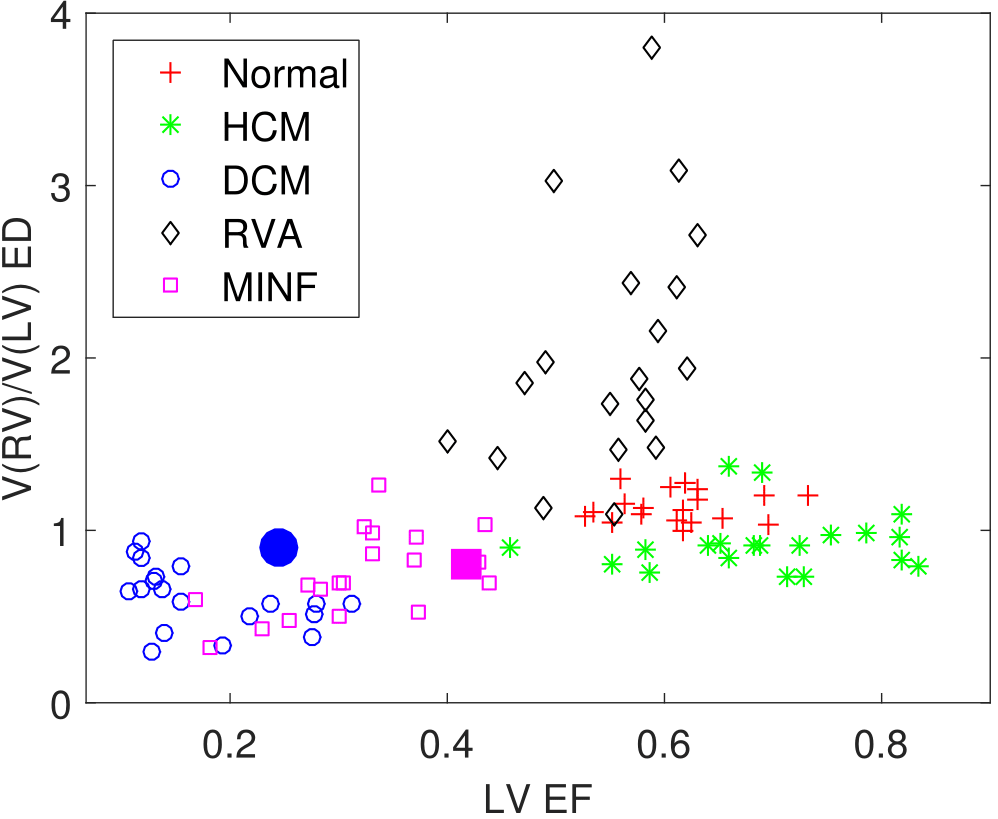}
\includegraphics[width=0.32\textwidth]{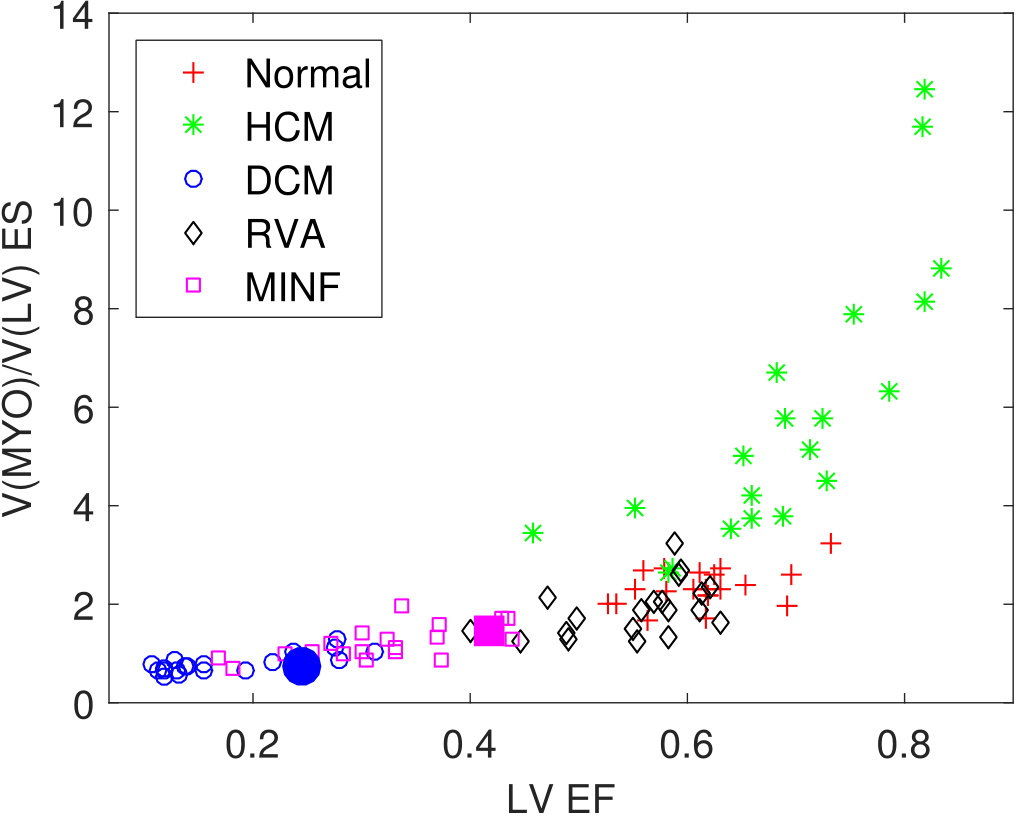}
\includegraphics[width=0.32\textwidth]{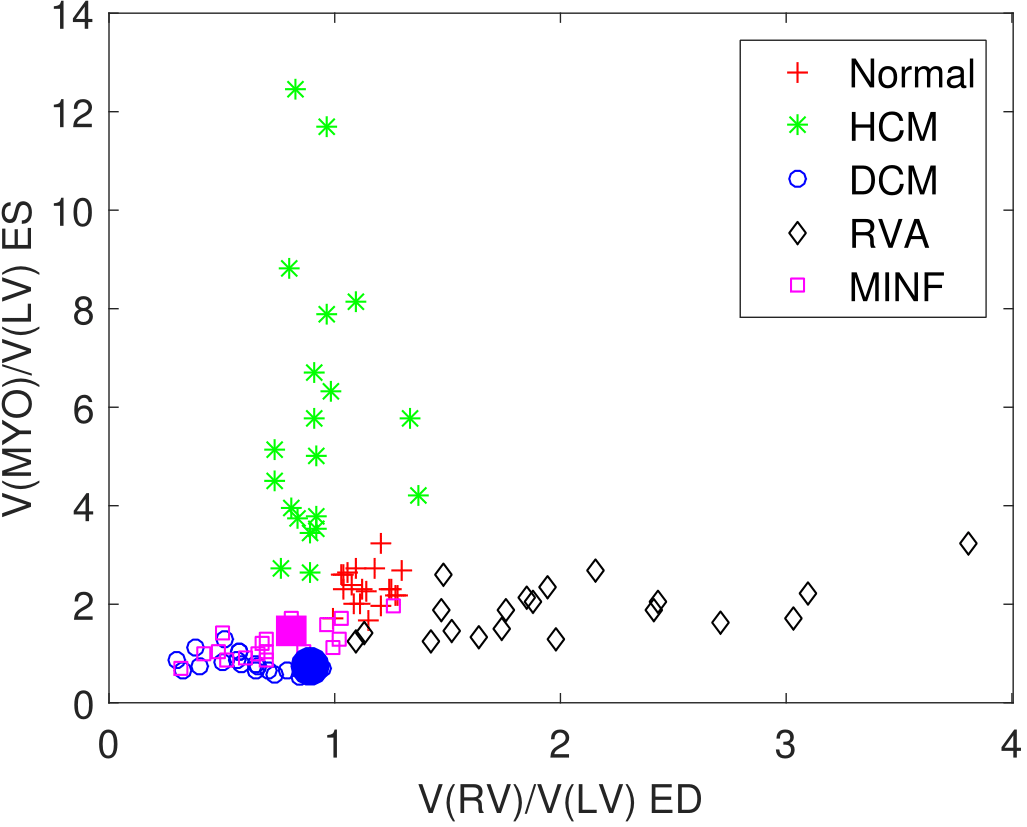}
\caption{Three most important features according to the Random Forest classifier: left ventricular ejection fraction (LV EF), volume ratio between right and left ventricle at ED (V[RV]/V[LV] ED), volume ratio between myocardium and left ventricle at ES (V[MYO]/V[LV] ES). Each point corresponds to a patient.}
\label{fig:pairplot}
\end{figure}

The three most important features as determined by the Random Forest were the left ventricular ejection fraction (LV EF), the ratio between right and left ventricular volume at ED (V[RV]/V[LV] ED), and the ratio between myocardial and left ventricular volume at ES (V[MYO]/V[LV] ES). Fig. \ref{fig:pairplot} shows the feature value distribution over the five different classes, showing several clear patterns. RVA patients generally have a large RV to LV volume ratio compared with normal patients. Patients with DCM and MINF have a reduced LV EF, while this value is higher for normal patients and patients with HCM. The myocardial volume is relatively small compared with the LV volume in patients with DCM, indicating thinning of the myocardium, but large in patients with HCM.

\begin{table}[tp!]
\centering
\caption{Agreement in diagnosis between the reference standard and automatic classification. Patients were classified as normal, dilated cardiomyopathy (DCM), hypertrophic cardiomyopathy (HCM), heart failure with infarction (MINF), or right ventricular abnormality (RVA). Overall classification accuracy was 91\%.}
\scriptsize
\begin{tabular}{lll|C{1.5cm}|C{1.5cm}|C{1.5cm}|C{1.5cm}|C{1.5cm}|c}
		 & \multicolumn{2}{c}{}&\multicolumn{5}{c}{Automatic}&\\	
		& \multicolumn{2}{c|}{}&Normal&DCM&HCM&MINF&RVA&\multicolumn{1}{c}{Total}\\
		\cline{3-8}
		&\multirow{5}{*}{\rotatebox[origin=c]{90}{Reference}}& Normal & \cellcolor[HTML]{C0C0C0}\textbf{20} & 0 & 0 & 0 & 0 & 20 \\
		\cline{3-8}
		&& DCM & 0 & \cellcolor[HTML]{C0C0C0}\textbf{18} & 0 & 2 & 0 & 20\\
		\cline{3-8}
		&& HCM  & 2 & 0 & \cellcolor[HTML]{C0C0C0}\textbf{18} & 0 & 0 & 20 \\
		\cline{3-8}
		&& MINF & 1 & 2 & 0 & \cellcolor[HTML]{C0C0C0}\textbf{17} & 0 & 20 \\		
		\cline{3-8}
		&& RVA & 2 & 0 & 0 & 0 & \cellcolor[HTML]{C0C0C0}\textbf{18} & 20 \\		
		\cline{3-8}
		&\multicolumn{1}{c}{} & \multicolumn{1}{l}{Total} & \multicolumn{1}{c}{25} & \multicolumn{1}{c}{20} & \multicolumn{1}{c}{18} & \multicolumn{1}{c}{19} & \multicolumn{1}{c}{18} & 100\\ 
	\end{tabular} 
\label{tab:diagnosis}
\end{table}

\begin{figure}[tp]
\centering
\subfloat[Patient 18 ED]{
\includegraphics[width=0.22\textwidth]{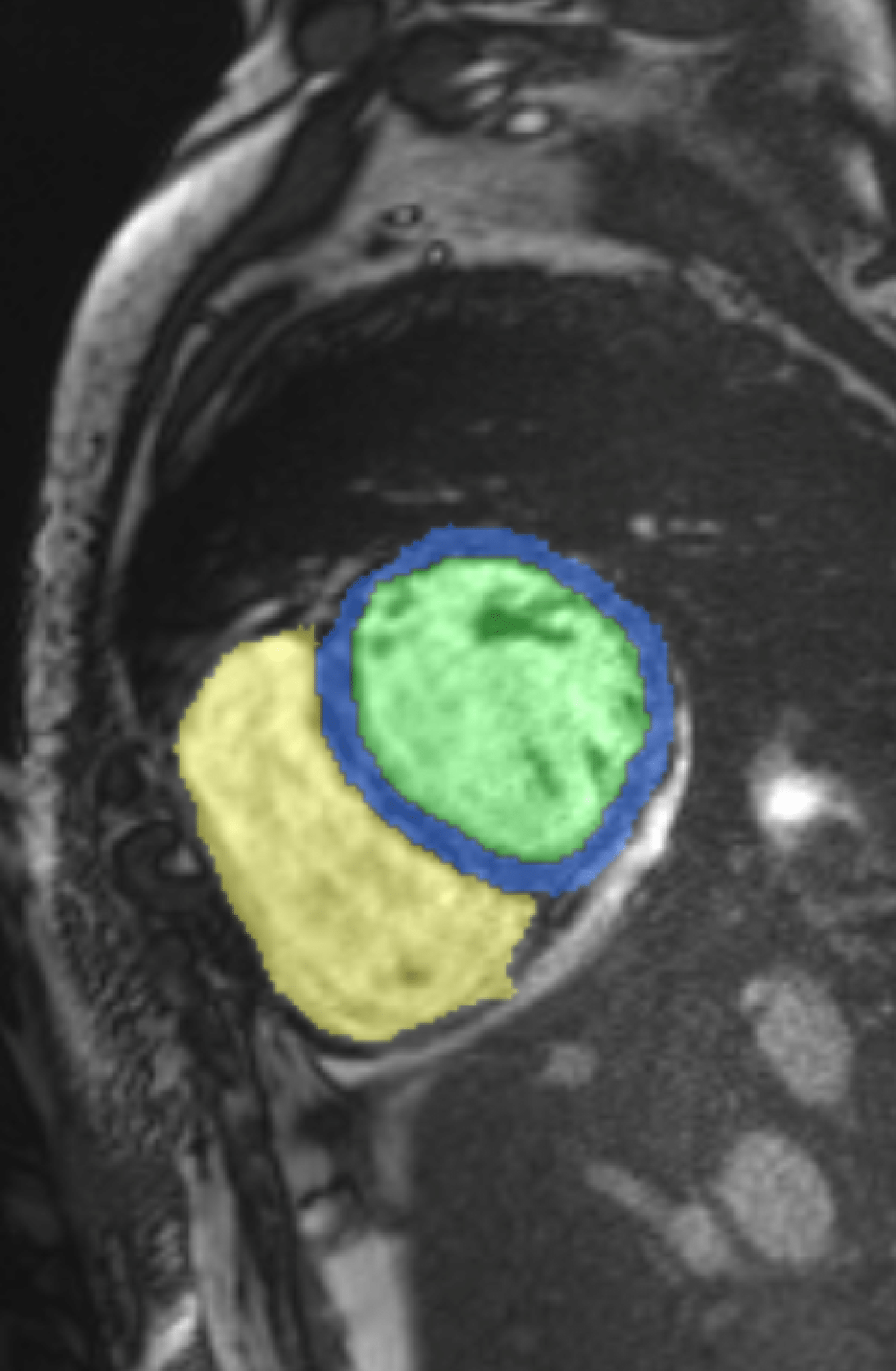}
\label{subfig:pat18ed}
}
\subfloat[Patient 18 ES]{
\includegraphics[width=0.22\textwidth]{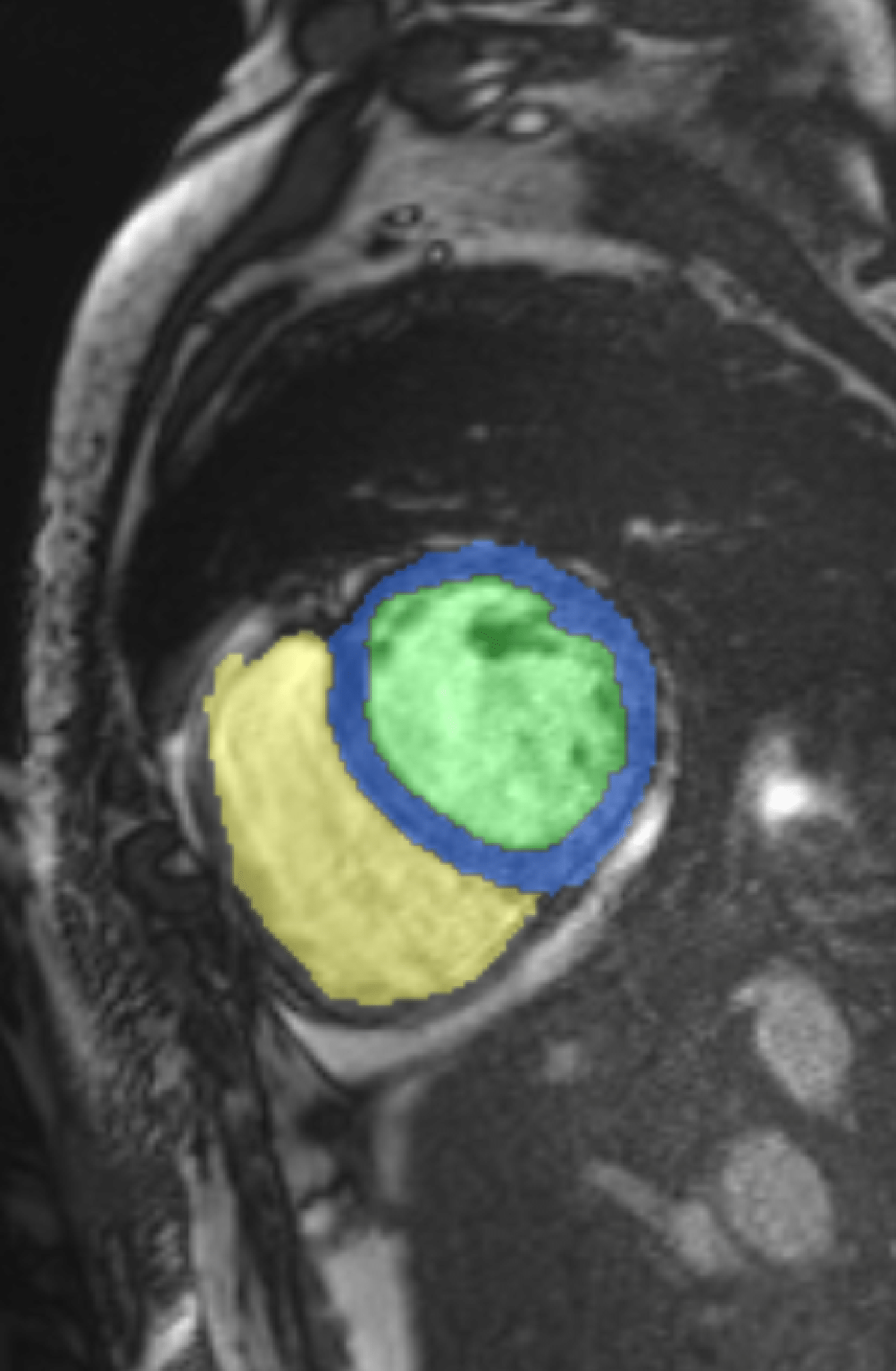}
\label{subfig:pat18es}
}
\subfloat[Patient 44 ED]{
\includegraphics[width=0.22\textwidth]{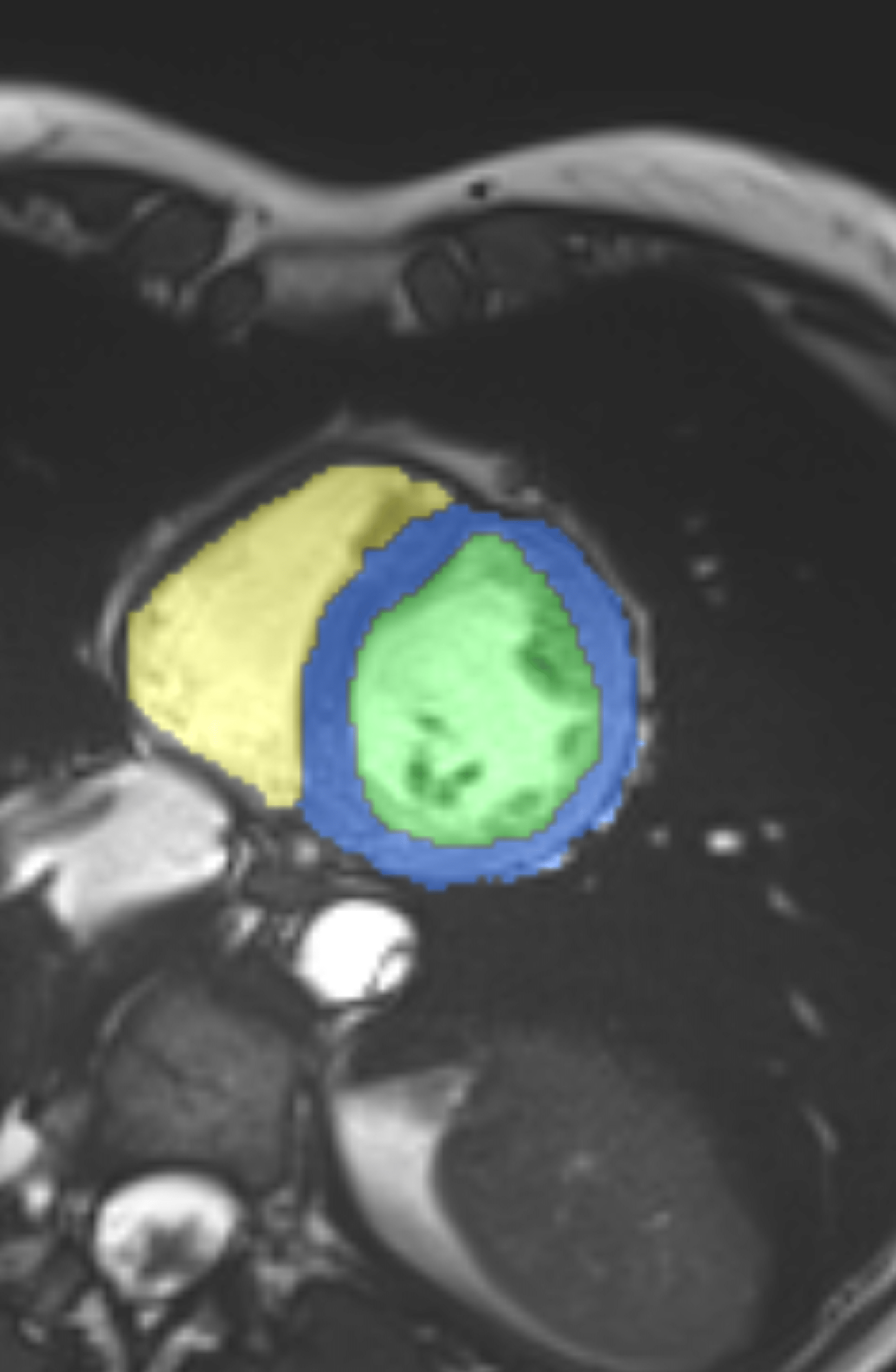}
\label{subfig:pat44ed}
}
\subfloat[Patient 44 ES]{
\includegraphics[width=0.22\textwidth]{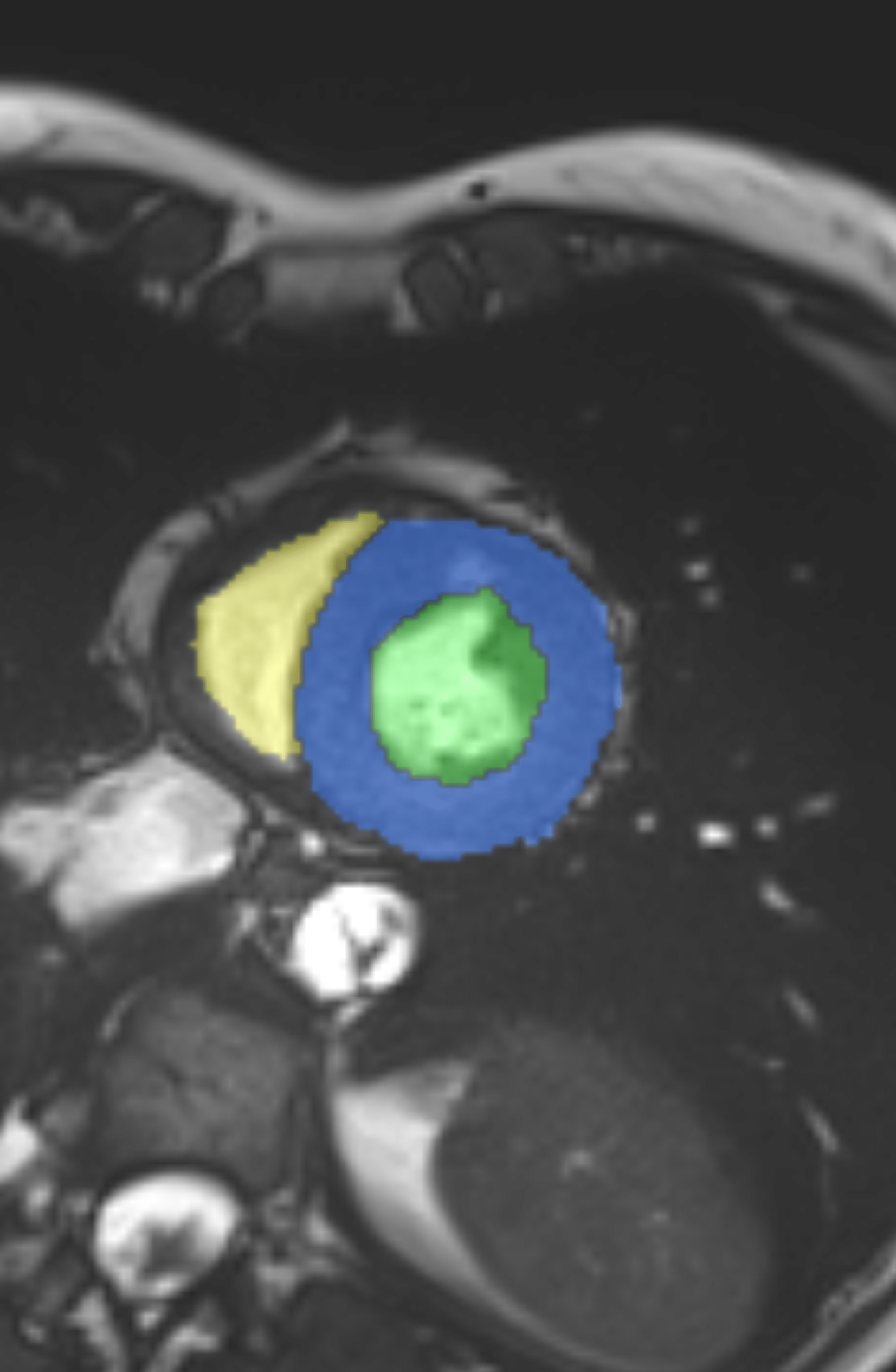}
\label{subfig:pat44es}
}
\caption{Two cases in which the classifier showed high uncertainty. The reference diagnosis for Patient 18 was DCM, but the patient was classified as MINF. The reference diagnosis for Patient 44 was MINF (LV EF 41.7\%), but the patient was classified as normal.}
\label{fig:uncertaincases}
\end{figure}

However, not all cases can be clearly separated using these features. Based on the entropy in the posterior probabilities provided by the Random Forest classifier, several patients with high classification uncertainty could be identified. Fig. \ref{fig:uncertaincases} shows automatically obtained segmentations in two such patients. Patient 18 (indicated by a blue circle in Fig. \ref{fig:pairplot}) was incorrectly diagnosed as MINF (classification probability $p=0.45$), while the reference diagnosis was DCM ($p=0.25$). In addition, there was a substantial probability for RVA ($p=0.21$). In this patient, LV EF and the ratio between RV and LV at ED were both relatively high compared with other DCM patients. 
Patient 44 (indicated by a magenta square in Fig. \ref{fig:pairplot}) was incorrectly diagnosed as normal ($p=0.44$), while the reference diagnosis was MINF ($p=0.31$). 
This patient had a high LV EF value of 41.7\% compared with other MINF patients.

Extraction of features based on segmentations and classification using the Random Forest classifier took around 1 s per patient.

\section{Discussion and Conclusion}
We have presented a method for fully automatic segmentation and diagnosis in cardiac cine MR images. The results show that automatically obtained segmentations of the left ventricle, right ventricle, and myocardium have good overlap with manual reference segmentations. Furthermore, based on these segmentations patients can be diagnosed with 91\% multi-class accuracy. 

While disease classification based on quantitative descriptors extracted from cine MR typically follows clinical guidelines, we have shown here that these guidelines can partially be captured in a Random Forest classifier. 
Furthermore, the posterior probability distribution of the classifier can be used to identify patients which cannot easily be assigned to a single disease. In future work, we will further investigate to what extent uncertainty of the classifier corresponds to uncertainty in the clinical diagnosis.

Deep learning methods have been shown to provide state-of-the-art results in a wide range of medical imaging problems. Here, we used deep learning for segmentation of the cine MR images. However, we opted for a more conventional Random Forest approach for patient classification because of the small training dataset at hand. A potential limitation of this two-stage approach is that errors in the segmentation stage may affect performance in the classification stage. However, a Bland-Altman comparison of quantitative indices derived from reference and automatic segmentations (Fig. \ref{fig:blandaltmans}) showed only very small bias values, comparable to interstudy and intraobserver differences in cardiac MR of normal healthy adults \cite{Mood15}. Moreover, patient classification using quantitative indices derived from the reference segmentations instead of the automatic segmentations resulted in only minor improvements (classification accuracy 92\% instead of 91\%) with a considerable increase in time and effort.

In the current study, we only used the ED and ES images. However, cine MR contains a whole sequence of images. In future work we will investigate whether inclusion of the complete sequence of images as input to the CNN could improve segmentation, and whether features derived from this sequence could provide additional value for disease diagnosis.

\bibliographystyle{splncs03}
\bibliography{wolterink}

\end{document}